\documentclass[final]{cvpr}

\usepackage{times}
\usepackage{epsfig}
\usepackage{graphicx}
\usepackage{amsmath}
\usepackage{amssymb}
\usepackage{multirow}

\usepackage{color, colortbl}
\usepackage{booktabs}
\usepackage{adjustbox}

\usepackage{tabu}

\usepackage{caption}

\definecolor{Gray}{gray}{0.92}

\usepackage{color}
\usepackage{xcolor}
\definecolor{citecolor}{HTML}{0071bc}
\usepackage[pagebackref=true,breaklinks=true,colorlinks,citecolor=citecolor,bookmarks=false]{hyperref}

\newlength\savewidth\newcommand\shline{\noalign{\global\savewidth\arrayrulewidth
  \global\arrayrulewidth 1pt}\hline\noalign{\global\arrayrulewidth\savewidth}}
  
\newcommand{\tablestyle}[2]{\setlength{\tabcolsep}{#1}\renewcommand{\arraystretch}{#2}\centering\footnotesize}
\usepackage{color}

\def\cvprPaperID{****} 
\def\confYear{CVPR 2021}

\begin{document}

\title{Synthesizing Long-Term 3D Human Motion and Interaction in 3D Scenes}

\author{Jiashun Wang\\UC San Diego
\and
Huazhe Xu\\UC Berkeley
\and
Jingwei Xu\\Shanghai Jiao Tong University
\and
Sifei Liu\\NVIDIA
\and
Xiaolong Wang\\UC San Diego

}

\twocolumn[{%
\renewcommand\twocolumn[1][]{#1}%
\maketitle
\vspace{-3.5em}
\begin{center}
    \centering
       \begin{tabular}{c}
    \hspace{-0.12in}
    \includegraphics[width=1\linewidth]{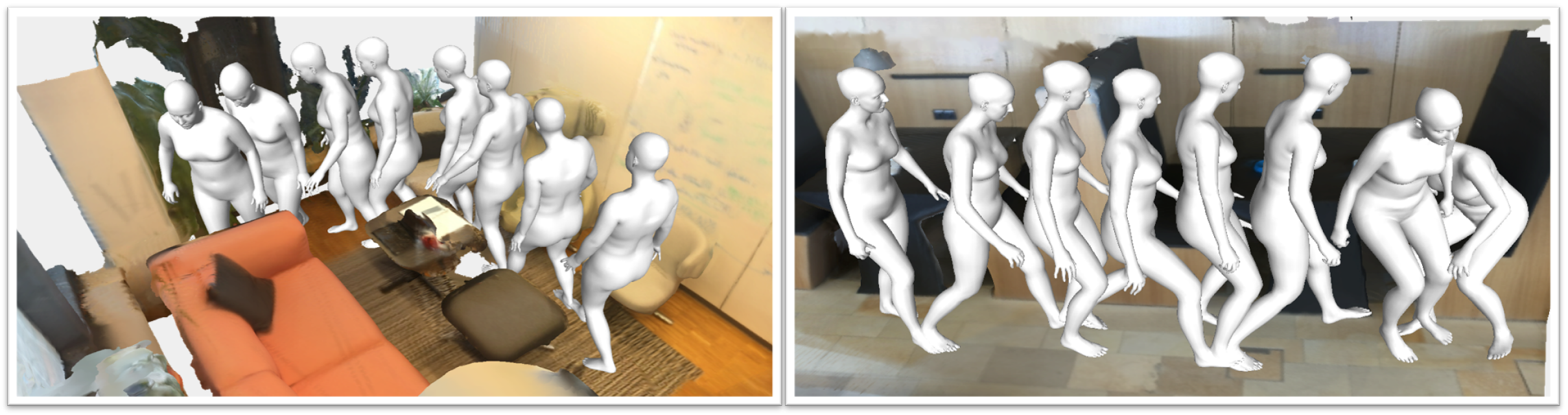}
        \end{tabular}
    \vspace{-0.12in}
    \captionof{figure}{Two examples of our generated long-term 3D motion in the 3D scene. \emph{Left}: A human walks around the furniture in the room; \emph{Right}: A human walks in the hallway and then sits down on the bench.  }
    \label{fig:teaser}
\end{center}%
}]
\begin{abstract}
Synthesizing 3D human motion plays an important role in many graphics applications as well as understanding human activity. While many efforts have been made on generating realistic and natural human motion, most approaches neglect the importance of modeling human-scene interactions and affordance. On the other hand, affordance reasoning (e.g., standing on the floor or sitting on the chair) has mainly been studied with static human pose and gestures, and it has rarely been addressed with human motion. In this paper, we propose to bridge human motion synthesis and scene affordance reasoning. We present a hierarchical generative framework to synthesize long-term 3D human motion conditioning on the 3D scene structure. Building on this framework, we further enforce multiple geometry constraints between the human mesh and scene point clouds via optimization to improve realistic synthesis. Our experiments show significant improvements over previous approaches on generating natural and physically plausible human motion in a scene.\footnote{Project page: \url{https://jiashunwang.github.io/Long-term-Motion-in-3D-Scenes}}
\end{abstract}

\vspace{-0.2in}
\section{Introduction}
\vspace{-0.02in}

Capturing and synthesizing realistic human motion in 3D scenes has played an essential role in various applications in virtual reality, video game animations and human-robot interactions. As shown in Fig.~\ref{fig:teaser}, given the 3D scenes, our goal is to generate long-term human motion and interaction in the scene, such as walking around the room avoiding collision with the furniture (left), as well as walking through the hallway, turning around and then sitting down (right). To achieve this, there are two main challenges on: (i) generating realistic motion in long-term; (ii) modeling human-scene interaction and affordance. 

Recent works have made substantial efforts on human motion synthesis, which generates visually appealing and natural pose sequences using optimization-based statistical models ~\cite{DBLP:journals/tog/XiaWCH15,DBLP:conf/siggraph/BrandH00}, or deep neural networks~\cite{DBLP:journals/tog/HoldenSK16,DBLP:journals/tog/HoldenKS17,xu2020hierarchical}. However, while focusing on realistic motion, these works rarely address the interactions between the human and the scene. On the other hand, a line of researches on 3D scene affordance~\cite{Wang_affordanceCVPR2017,3d-affordance,zhang2020generating} has studied the ``opportunities for interactions''~\cite{gibson79} in the scene. For example, Wang \textit{et al.}~\cite{Wang_affordanceCVPR2017} propose to learn to predict human skeletons from an empty scene by training with a large-scale sitcom dataset. While focusing on the scene context, these approaches are only able to generate a single static human pose.

In this paper, we intend to bridge human motion synthesis and affordance learning. We consider a novel problem setting: \emph{Given the start and the end positions far away in a 3D scene, synthesize the human motion moving in between.} To generate long-term motion in the scene, instead of synthesizing a long route of poses at one time, we introduce a 2-level hierarchical framework: (i) we first set several sub-goal positions between the start and the end locations.  We predict the human pose for each sub-goal, start and end positions, conditioning on the 3D scene context. (ii) we synthesize the short-term human motion between every two sub-goals, using the predicted poses on the sub-goals as well as the 3D scene as inputs. The short-term motion will be then connected together for the final long-term motion synthesis. 

We model the interaction between human and the scene in both stages of our framework. Instead of using human skeletons, we emphasize that we adopt the differentiable SMPL-X~\cite{pavlakos2019expressive} model for representing both the shape and the pose of the human, which allows more flexible geometry constraints and more realistic modeling of contacts. Specifically, in the first stage, given a single sub-goal in a 3D scene, we utilize a Conditional Variational Autoencoder (CVAE)~\cite{sohn2015learning} to generate the SMPL-X parameters. In the second stage for short-term motion generation, we use a bi-directional LSTM~\cite{DBLP:journals/neco/HochreiterS97} which takes the start-end human SMPL-X representations and 3D scene representation as inputs and generates a sequence of human bodies represented by SMPL-X. Besides training the deep models in a data-driven manner for motion synthesis, we also perform explicit geometry reasoning between the human mesh and 3D scene point clouds by optimization. Our optimization approach considers both the naturalness of motion and the physical collisions with the environment. By unifying both learning-based and optimization-based techniques, we are able to synthesize realistic human motion in long-term. 

We perform our experiments on both the PROX~\cite{hassan2019resolving} and the MP3D~\cite{chang2017matterport3d} 3D environments. By considering 3D scene affordance and structural constraints in motion synthesis, we qualitatively show realistic and physically plausible human motion generation results. We also show large advantages quantitatively against state-of-the-art motion and human pose generation approaches, using multiple metrics and human evaluation. 

Our contributions in this paper include: (i) A hierarchical learning framework for motion synthesis considering both the realism of the motion and the affordance of the scene; (ii) An optimization process to explicitly improve the synthesized human poses; (iii) state-of-the-art motion synthesis results on various 3D environments.

\section{Related work}

\textbf{Affordance learning.} Learning scene affordance has captured a lot attention in recent years~\cite{fouhey2014people,wang2019geometric,tan2018and,ouyang2018pedestrian,chuang2018learning,kim2014shape2pose,delaitre2012scene,gupta20113d,koppula2013learning,3d-affordance,savva2016pigraphs,zhu2016inferring,zhu2014reasoning,monszpart2019imapper,savva2014scenegrok,chen2019holistic++,zhang2020generating,zhangplace,cao2020long}. One paradigm to study the scene affordance is to understand how to put a human skeleton  in a scene~\cite{tan2018and, 3d-affordance,zhang2020generating,zhangplace,tan2018and,ouyang2018pedestrian}. For example, Tan \textit{et al.}~\cite{tan2018and} predict where to put a human in a given image and search for a person that fits the scene from a database. Li \textit{et al.}~\cite{3d-affordance} introduce a 3D pose generative model to predict physically feasible human poses in a given scene.  Recently, due to more refined needs and the development of 3d human representations~\cite{loper2015smpl, romero2017embodied, pavlakos2019expressive}, more researches start to study how to place a human body shape in a 3d scene instead of skeleton. For example, Zhang \textit{et al.}~\cite{zhang2020generating} use a two-stage CVAE~\cite{sohn2015learning} to generate plausible 3D human bodies that are posed naturally in 3D scene. However, most studies only focus on the single static body generation and they have barely addressed the problem of generating physically plausible motion in the scene. 

\textbf{Human dynamics prediction.} Our work is related to the research in modeling and predicting the human dynamics. Both research in trajectory prediction~\cite{helbing1995social,treuille2006continuum,kitani2012activity,alahi2016social,hochreiter1997long,gupta2018social,sadeghian2019sophie,alahi2014socially,makansi2019overcoming,tai2018socially,cao2020long} and pose prediction~\cite{villegas2017learning,zhao2018learning,walker2017pose,jain2016structural,holden2015learning,li2018convolutional,fragkiadaki2015recurrent,hernandez2019human,mao2019learning,pavllo20193d} are raising a lot of attention in recent years. Instead of isolating the environments from prediction, researchers have studied on using 3D information or bird-eye view image to predict future human dynamics~\cite{gupta2018social,sadeghian2019sophie,alahi2014socially,makansi2019overcoming,tai2018socially}. To study on how the surroundings would influence the human dynamics, Helbing \textit{et al.}~\cite{helbing1995social} use physical forces to model social-scene interaction for pedestrian dynamics. Cao \textit{et al.}~\cite{cao2020long} consider scene context when predicting the goal, path and poses of human movement given the scene image and past 2D pose histories as inputs. While these studies make a great contribution to the human dynamics and skeleton-based pose prediction, they pay little attention to the physical and geometric interaction between the human and the scene for realistic generation. On the contrary, our paper focuses on synthesizing natural human pose and shape when interacting with the scene. We also want to emphasize we are not performing future prediction since our goal is given in the task. 
\begin{figure*}[!t]
\begin{center}
\hspace{-0.2cm}
\includegraphics[width=1\linewidth]{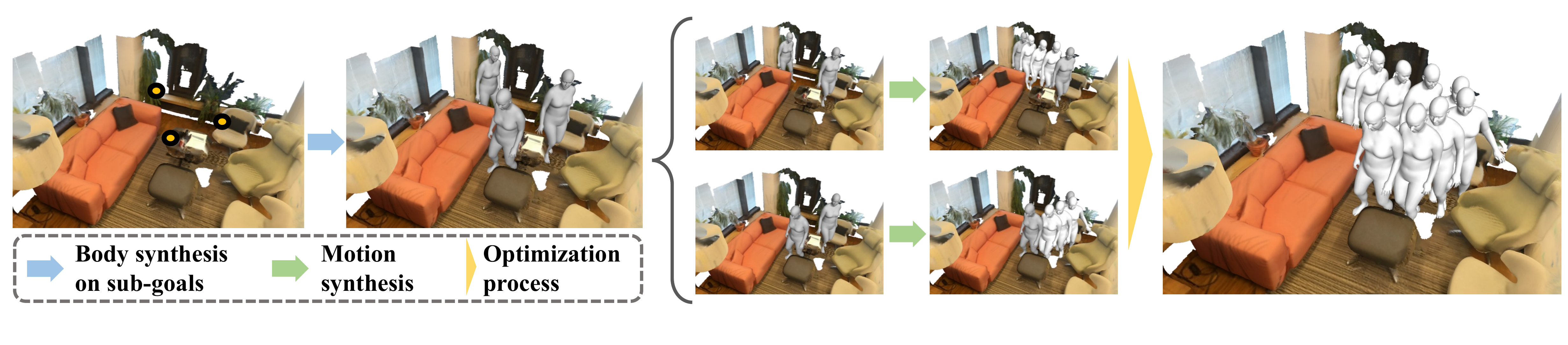}
\end{center}
\vspace{-0.4in}
\caption{\textbf{Framework of our long-term motion synthesis.} We first generate the sub-goal bodies with the given $\{\beta,t,r\}$ as inputs. Sub-goal bodies are in gray color. Then we divide it into several short-term start/end pairs and synthesize short-term motion each. Finally we use an optimization process to connect all these short-term motion to a long-term motion. Generated motion is in white.}
\vspace{-0.18in}
\label{fig:pipline}
\end{figure*}

\textbf{Motion synthesis.} The main focus of our work is on motion synthesis, which has been long standing problem in computer graphics and vision~\cite{DBLP:conf/siggraph/BrandH00,DBLP:conf/sca/KovarG03,DBLP:conf/sca/ParkSS02,DBLP:journals/jvca/TanT12,DBLP:journals/tog/HoldenSK16,DBLP:journals/tog/LiWS02,pavllo2018quaternet,pavlovic2000learning,xu2020hierarchical}. For example, Kovar \textit{et al.}~\cite{DBLP:conf/sca/KovarG03} introduce a novel data structure called a registration curve that expands the class of motions that can be successfully blended without manual input. Pavllo \textit{et al.}~\cite{pavllo2018quaternet} represent rotations with quaternion and uses a sequential network and a novel loss function to perform forward kinematics on a skeleton to penalize absolute position errors instead of angle errors. There are also studies ~\cite{DBLP:conf/icml/UrtasunFGPDL08,DBLP:journals/tii/XiaSLH19,harvey2020robust,liu2002synthesis} focusing on synthesizing intermediate states between the given key frames. But these works can only synthesize a transient motion with a small position change. Xu \textit{et al.}~\cite{xu2020hierarchical} proposes a hierarchical way to generate long-term motion by using a memory bank to retrieve short-term motion references and a bi-directional interpolation to connect the short-term motions. Several studies~\cite{starke2019neural,heess2017emergence,clegg2018learning,ling2020character} have considered motion and the environment. However, these methods either rely on predefined objects and primitive motion or use simple hand-crafted environments.
Our method does not use these assumptions and can generate motion directly from the realistic scene point cloud. This allows our model to not only generalize beyond predefined motion, but also generalize better to unseen 3D environments. To better model detailed human-scene interactions, we use the representation of human mesh instead of human skeleton during motion synthesis.






\section{Method}
\subsection{Overview}

\textbf{Representation.} We denote scene mesh as $S=(v^{s},f^{s})$ where $v^{s}$ represents the vertices  and $f^{s}$ represents the faces. Instead of using skeleton-based representation, we use modified sequential SMPL-X parameters~\cite{pavlakos2019expressive} to represent human bodies. Concretely, we define  $M_{i}=\mathcal{M}(t_{i},r_{i},\beta ,p_{i},h_{i})$, where $t \in \mathbb{R} ^{3}$ is the translation, 6d continuous rotation
$r \in \mathbb{R} ^{6}$~\cite{zhou2019continuity} is used to replace the original global orientation in SMPL-X,
$\beta \in \mathbb{R} ^{10}$ is the body shape,
$p \in \mathbb{R} ^{32}$ represents the body pose~\cite{pavlakos2019expressive} and $h \in \mathbb{R} ^{24}$ represents the hand pose.

\textbf{Problem definition.} We design a two-level hierarchical framework to generate long-term 3D motion in the 3D  scene. Our method takes the inputs of the start, end and sub-goal positions and orientations. The sub-goals divide the higher-level long trajectory into lower-level short paths. We first generate the human bodies on these given positions with given shape $\beta$, using a Conditional Variational Autoencoder (CVAE)~\cite{sohn2015learning}. We give $\beta$ to control the body shape of the motion with more diversity. Given the generated human bodies on each sub-goal, we aim to generate plausible motion between every two sub-goals, which will then be connected together to a long-term motion. 

Concretely, we propose a motion synthesis network to generate short-term motion. Given the start body mesh $M_{0}=\mathcal{M}(t_{0},r_{0},\beta,p_{0},h_{0})$, the end body mesh $M_{k}=\mathcal{M}(t_{k},r_{k},\beta ,p_{k},h_{k})$ estimated by the CVAE and the scene mesh $S$ as inputs, the motion synthesis network will generate a motion sequence $M_{1:k-1}$ between the start and end bodies. Multiple short-term motion sequence $M_{1:k-1}$ will be connected together to form a long-term motion. We assume the length of each short-term motion is $k+1$ steps. Finally, we adopt a geometric optimization process to further enforce realistic and physically plausible synthesis. 

The benefits of our two-level generation approach lie in two folds: (i) First generating the poses on the sub-goals and then generating the motion in between allows the short-term motion easier to connect, since the ending pose of one sequence will be the same as the starting pose of the next sequence. At the same time, generating a short-term motion reduces the uncertainty on model prediction; (ii) Using sub-goals allows diverse long-term motion generation. By sampling different latent variables in CVAE, we can generate diverse human poses on the sub-goals, which leads to diverse motion in the long-term trajectory.

\begin{figure*}[t]
\begin{center}
\includegraphics[width=0.95\linewidth]{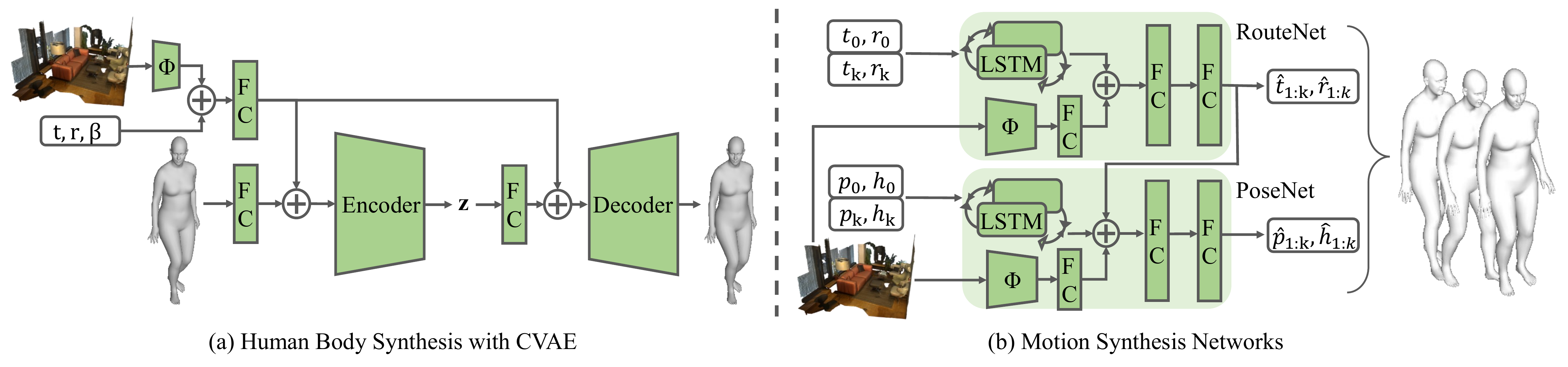}
\end{center}
\vspace{-0.7cm}
\caption{\textbf{Network architectures.} (a) shows the architecture of our static human body synthesis network. (b) is the architecture of our motion synthesis networks. $\bigoplus$ means concatenation.}
\vspace{-0.2in}
\label{fig:network}
\end{figure*}

\subsection{Static Human Body Synthesis on Sub-Goals}

We propose to use a Conditional Variational Autoencoder (CVAE) for generating bodies on each sub-goal, given the inputs of $\{\beta,t,r\}$ presented in the sub-goal and the scene point cloud $v^{s}$, as shown in Fig.~\ref{fig:network} (a). During training time, we first extract the feature for the scene point cloud as  ${F}^{s} = \Phi (v^s)$ using a PointNet $\Phi$~\cite{qi2017pointnet}. We concatenate it with the shape, location and orientation inputs $\{\beta,t,r\}$ and forward them to a fully connected layer to obtain the integrated feature ${F}^{hs}$. This feature is the conditional feature for the CVAE.  The CVAE is presented with an encoder and a decoder as follows. 

\textbf{Encoder.} We forward the human body parameters $M_{0}$ to two residual blocks containing two fully connected layers each. The output is then concatenated with ${F}^{hs}$, followed by two fully connected layers to predict the mean $\mu \in \mathbb{R}^{32}$ and variance $\sigma^2 \in \mathbb{R}^{32}$ of a Gaussian distribution $Q(z | \mu, \sigma^2)$. We sample the latent code  $z$ from this distribution as one of the decoder inputs. 

\textbf{Decoder.} We concatenate the latent code $z$ with the conditional feature ${F}^{hs}$ as the input for the decoder, which is another two residual blocks containing two fully connected layers each. The output of the decoder predicts the desired human body parameters, which is trained computing the reconstruction loss against the ground-truth human mesh $M_{0}$. Following~\cite{kingma2013auto}, the training objective also includes maximizing the KL-Divergence between $Q(z | \mu, \sigma^2)$ and a standard Gaussian distribution $\mathcal{N}(0, I^2)$. 

During inference, only the decoder is adopted. We sample the latent code from $\mathcal{N}(0, I^2)$ and concatenate it with ${F}^{hs}$ as inputs. By sampling different $z$, we can generate different human bodies in the same sub-goal position.  

\textbf{Implementation details for PointNet.} We modify the PointNet architecture for affordance prediction. We remove the transformation architecture to better fit our task of generating a body mesh in a specified location, since applying the transformation to the scene may cause the mismatch of human and scene coordinates. We pretrain the PointNet using the  S3DIS~\cite{armeni20163d} dataset with a segmentation task. We then take the encoder of the PointNet as $\Phi$ and the output is a 256-dimension feature for point cloud representation. 

To train the CVAE for static human body synthesis, we adopt the standard reconstruction loss and the KL divergence loss proposed in~\cite{sohn2015learning}, followed by the contact and collision constraints proposed in~\cite{zhang2020generating}. 

\subsection{Motion Synthesis Framework}
We propose two  sequentially connected networks in the motion synthesis framework: A RouteNet $\mathcal{R}$ for predicting the locations and orientations of the route between two sub-goals, and a PoseNet $\mathcal{P}$ for predicting the human body pose on each locations of the route. 

As shown in the top row of Fig.~\ref{fig:network} (b), the RouteNet $\mathcal{R}$ takes the start $\{t_{0},r_{0}\}$ and the end $\{t_{k},r_{k}\}$ locations and orientations, as well as the scene point cloud $v_{s}$ as inputs, and generates the route in between as, 
\begin{equation}
\widehat{t}_{1:k-1},\widehat{r}_{1:k-1}=\mathcal{R}(t_{0},r_{0},t_{k},r_{k}, v^{s})
\end{equation}
where $\{\widehat{t}_{1:k-1},\widehat{r}_{1:k-1}\}$ represents the route locations and orientations from step $1$ to $k-1$. Concretely, to extract the feature for of the scene, we utilize a PointNet given $v_{s}$ as inputs. Note that the PointNet feature extractor is not shared with the CVAE in the previous section. We denote the point cloud feature as $F^{rs}$. To connect the start and end locations, we utilize a bi-directional LSTM~\cite{DBLP:journals/neco/HochreiterS97} which takes the start $\{t_{0},r_{0}\}$ and the end $\{t_{k},r_{k}\}$ as inputs and outputs the features for each time step in between. We concatenate the features from all time steps as well as the point cloud feature $F^{rs}$ together. The output is then forwarded to two fully connected layers for predicting the route $\{\widehat{t}_{1:k-1},\widehat{r}_{1:k-1}\}$. 

We illustrate the PoseNet $\mathcal{P}$ as the bottom row of Fig.~\ref{fig:network} (b). It takes the start pose $\{p_{0},h_{0}\}$, the end pose $\{p_{k},h_{k}\}$,  point cloud $v^{s}$ and the predicted route $\{\widehat{t}_{1:k-1},\widehat{r}_{1:k-1}\}$ from the RouteNet as inputs. The outputs of the PoseNet are the pose parameters for the input route as, 
\begin{equation}
\begin{split}
\widehat{p}_{1:k-1},\widehat{h}_{1:k-1} = \mathcal{P}  (p_{0},h_{0},p_{k},h_{k},\widehat{t}_{1:k-1},\widehat{r}_{1:k-1},v^s)
\end{split}
\end{equation}
where $\{\widehat{p}_{1:k-1},\widehat{h}_{1:k-1}\}$ are the body pose and hand pose parameters from step $1$ to $k-1$. Similar to the RouteNet, we use another PointNet to extract the feature for $v^s$ as $F^{ps}$. The start and end pose parameters $\{p_{0},h_{0}\}$ and $\{p_{k},h_{k}\}$ are also forwarded to another bi-directional LSTM, which predicts the pose features between the two locations. We concatenate the pose features from step $1$ to $k-1$ and $F^{ps}$ together. We then forward the feature to two fully connected layers for generating the a sequence of pose parameters $\{\widehat{p}_{1:k-1},\widehat{h}_{1:k-1}\}$. Finally, we combine the predicted SMPL-X~\cite{pavlakos2019expressive} parameters $\{t_{1:k}, r_{1:k}, p_{1:k}, h_{1:k}\}$ and the given $\beta$ to generate the mesh sequence $M_{1:k}$.

We compute the training losses for both the RouteNet and the PoseNet with the $\ell_{1}$ distance between the predictions and the ground-truth location and pose parameters. Specifically, we define the RouteNet loss $\mathcal{L}_{route}$ and the PoseNet loss $\mathcal{L}_{pose}$ as, 
\begin{equation}
\mathcal{L}_{route}=\lambda_{t}\sum ^{k-1}_{i=1}|\widehat{t}_{i}-t_{i} |+\lambda_{r}\sum ^{k-1}_{i=1}|\widehat{r}_{i}-r_{i} |
\end{equation}
\begin{equation}
\mathcal{L}_{pose}=\lambda_{p}\sum ^{k-1}_{i=1}|\widehat{p}_{i}-p_{i} |+\lambda_{h}\sum ^{k-1}_{i=1}|\widehat{h}_{i}-h_{i} |
\end{equation}
where $\lambda_{t}$, $\lambda_{r}$, $\lambda_{p}$ and $\lambda_{h}$ are constant coefficients to balance the loss. During training, we train the RouteNet first and then fix its weight for training the PoseNet.

\begin{figure}[]
\begin{center}
\vspace{-0.05in}
\includegraphics[width=0.88\linewidth]{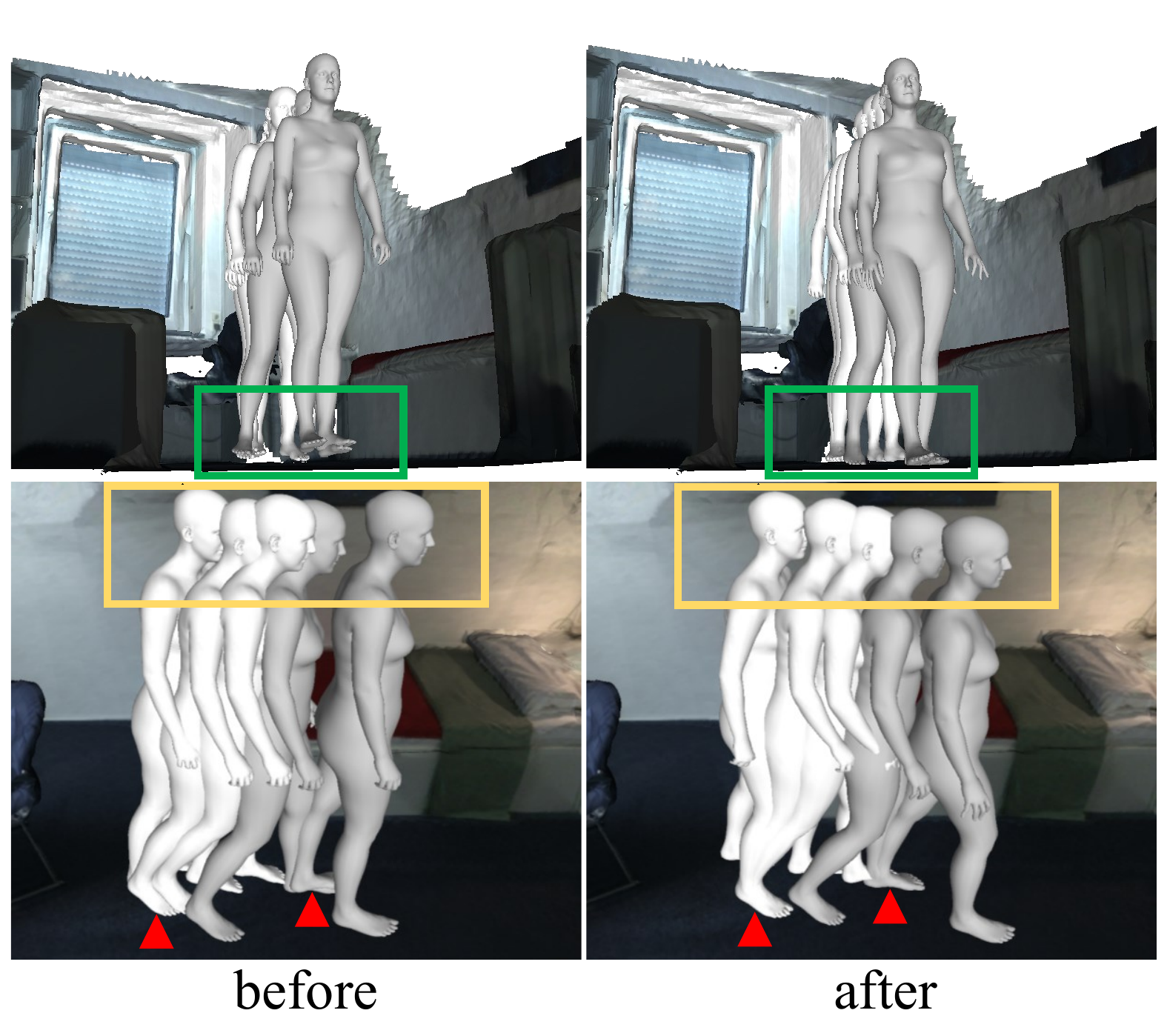}
\end{center}
\vspace{-0.3in}
  \caption{An example of human motion before and after optimization in two views. The green box shows that after optimization, the motion sequence is contacting with the floor. The red arrow points at the pivot foot. White human bodies are from the first sub-sequence, gray bodies are from the second, it can be seen that in each sub-sequence, the pivot foot is more stable after optimization. Yellow boxes show the effectiveness of vertices smoothing.}
\vspace{-0.22in}
\label{fig:optimization}
\end{figure}

\subsection{Optimization}
We perform optimization with geometric and physics constraints to improve the generation results from the motion synthesis networks and help connect the short-term motions to a long-term motion at the same time. We will introduce three types of constraints in the following, including the constraint on the foot, contact constraints and the smoothness of the motion. For simplicity, we will take the generated mesh $M_{1:k-1}$ as an example for explaining our approach. While in our experiments, we apply the optimization approach to the whole long-term sequence.

\textbf{Inputs and variables.} The scene point cloud is provided to calculate the contact and we have the signed distance field~\cite{hassan2019resolving} of the scene mesh surface to calculate the collision. All the SMPL-X parameters are given and our goal is to optimize the translation $t$, global orientation $r$, body pose $p$ and also hand pose $h$.

\textbf{Foot location constraints.} While most previous works~\cite{li2017auto, xu2020hierarchical,aberman2020unpaired} assume the floor is flat and also have information on which foot should be fixed on the floor at each time step, our approach does not make any of such assumptions. This allows the generated motion to be more diverse and natural \eg sitting down or jumping. However, we still need to constrain the stableness of the foot motion.

As the human is moving, in each moment, we assume one foot is stable and the other is moving. Thus we aim to separate the motion into multiple sub-sequences, where each sub-sequence has the same foot stable and this foot switches between the sub-sequences. We get the sub-sequences utilizing the nearer foot between two frames from networks' outputs: nearer one should be the stable one. We compute the average location of the stable foot in each sub-sequences, and we encourage the foot to be close to this average location since it should stand still. We use the $\ell_{2}$ distance for computing this error as $E_{foot}$ for adjusting the foot location.  

\textbf{Environment constraints.} We also consider the physical plausibility between the human and the environment during optimization. The constraint design here is motivated by~\cite{zhang2020generating,hassan2019resolving}. On one hand, we constrain the human mesh to avoid collision with the scene; On the other hand, we also encourage the human mesh to get close to the scene for physical support. 

For collision constraint, we utilize the negative signed distance field of the scene $\Psi^{-}_{s}(\cdot)$, where we constrain the human mesh from intersecting with the scene 3D surfaces, we can represent the collision error as,
\begin{equation}
E_{col}=\sum^{k-1}_{i=1}\mathbb{E}(|\Psi^{-}_{s}(M_{i})|)
\end{equation}
where $|\Psi^{-}_{s}(M_{i})|$ computes the negative signed distance values of the body vertices, $\mathbb{E}$ represents an average function. The goal is to minimize the loss so that the body stays on the positive level-set of the signed distance field. 

We also encourage the human body to contact the scene. Our goal is to minimize $E_{cont}$ as below,
\begin{equation}
E_{cont}=\sum^{k-1}_{i=1}\sum_{v' \in v^{c}_i}\min_{v'' \in v^{s}}\rho(|v'-v''|)
\end{equation}
where $v^{c}_i$ denotes a set of the predefined~\cite{hassan2019resolving} body vertices which are encouraged to contact with the scene vertices. The Geman-McClure error function $\rho(\cdot)$ is used to down-weight the scene vertices which are far from the human. 

\textbf{Motion smoothness.}
We encourage the human mesh nearby in time to be smooth. To achieve this, we define the smoothness constraint to be minimizing the $\ell_{2}$ distance between the mesh vertices as, 
\begin{equation}
E_{smooth}=\sum^{k-2}_{i=1}\left\|v^{M_{i}}-v^{M_{i+1}} \right\|_{2} 
\end{equation}
where $v^{M_{i}}$ is the vertices of body $M_{i}$, and we perform this constraint between every two consecutive steps. 

During optimization, we combine all the three error terms together as $\lambda_{foot} E_{foot}+\lambda_{col} E_{col}+\lambda_{cont} E_{cont}+\lambda_{smooth}E_{smooth}$. $\lambda_{foot}$, $\lambda_{col}$, $\lambda_{cont}$ and $\lambda_{smooth}$ are constant coefficients. We perform gradient descent with Adam~\cite{kingma2014adam} to optimize the predicted mesh parameters directly. We show the effects of our optimization in Fig.~\ref{fig:optimization}.



\vspace{-0.05in}
\begin{figure*}[!t]
\begin{center}
\includegraphics[width=0.95\linewidth]{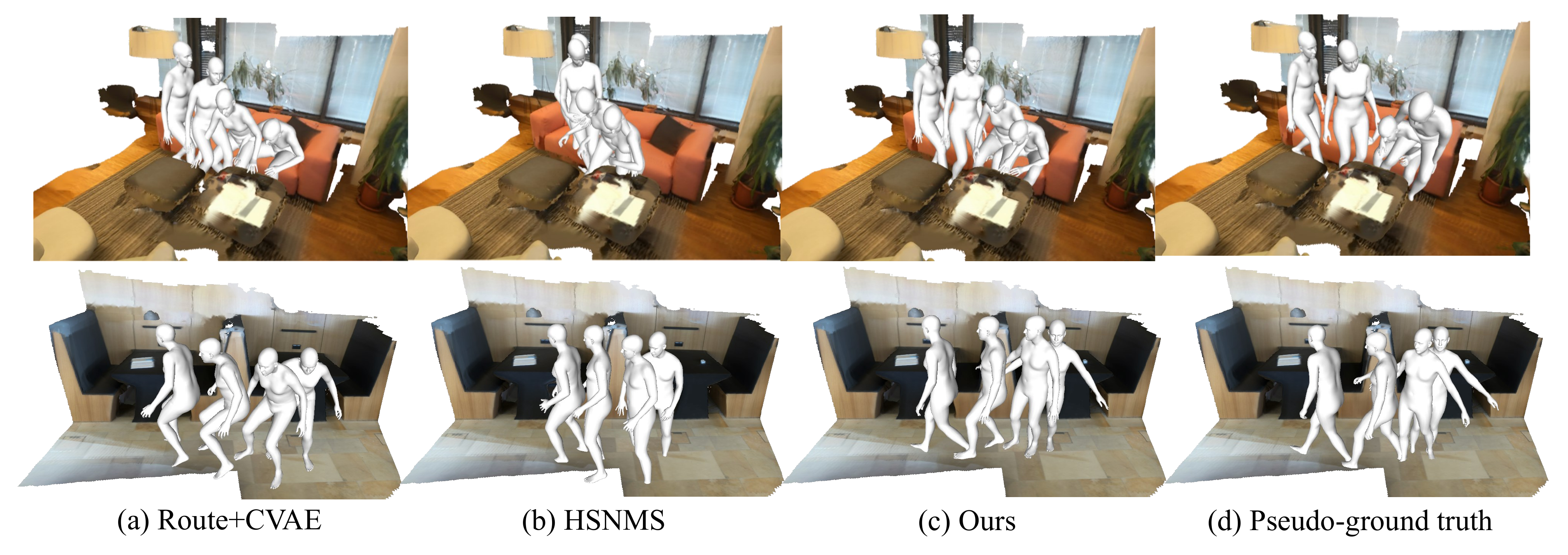}
\end{center}
\vspace{-0.32in}
   \caption{Comparisons on generating 2-second motion given the same inputs: (a) the result of Route+CVAE~\cite{zhang2020generating}; (b) the result of HSNMS~\cite{xu2020hierarchical}; (c) our result; (d) the pseudo-ground truth. We show that our results have more natural motions and a more consistent sitting down motion. (b) and (d) have shown cases of foot inside the ground. The poses in (a) are not natural.}
\label{fig:2_seconds_compare}
\end{figure*}

\begin{figure*}[t]
\vspace{-0.2in}
\begin{center}
\includegraphics[width=0.95\linewidth]{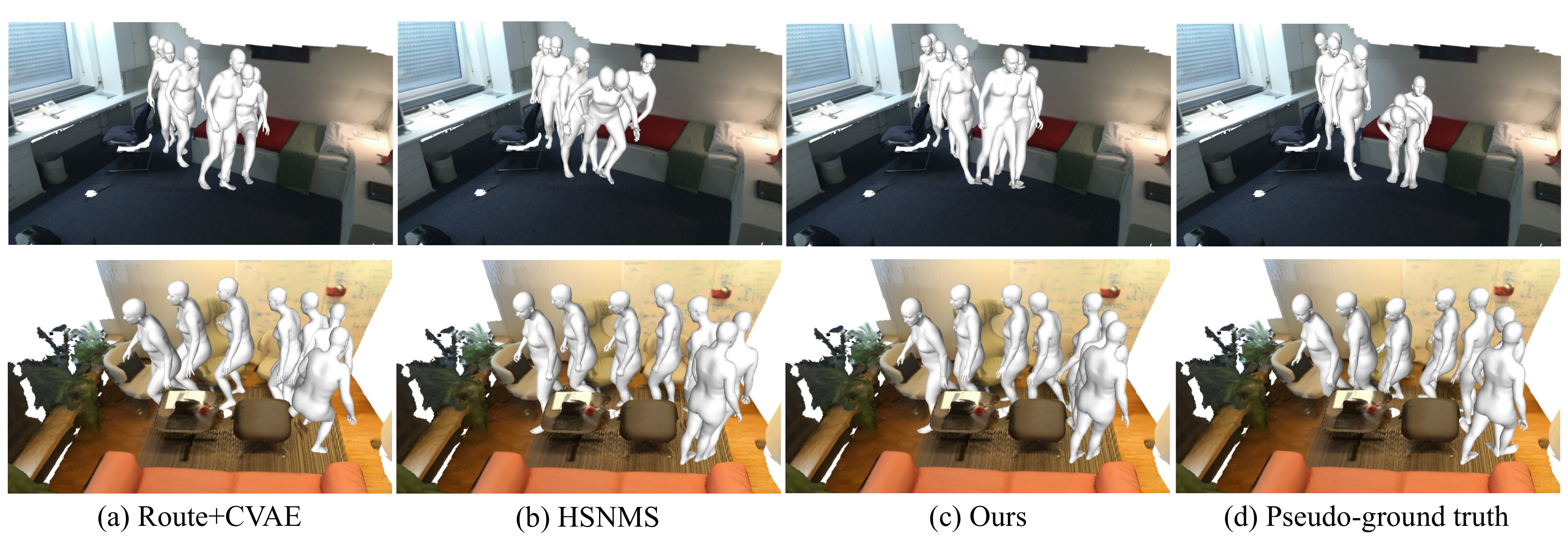}
\end{center}
\vspace{-0.33in}
   \caption{Comparisons on generating 4-second motion given the same inputs over different methods: While our method shows natural and physically plausible motions, (b) has the problem in generating suitable poses in the first row, and the motion in (a) does not look natural.}
\label{fig:4_seconds_compare}
\vspace{-0.2in}
\end{figure*}

\section{Experiments}
\subsection{Implementation details}
We use Adam~\cite{kingma2014adam} as the optimizer for our networks and also the optimization process. To train our CVAE generation network, we use 0.001 as the learning rate (lr) with a batch size of 16 and training epochs of 40. For RouteNet the lr is 0.001. We train 20 epochs with a batch size of 32. We train PoseNet with 0.001 lr and a batch size of 16 with 20 epochs. We set $\lambda_{t}=\lambda_{r}=\lambda_{p}=1$ and $\lambda_{h}$ is chosen as 0.1. Our short-term motion lasts for 2 seconds (30 fps), which means $k=61$ in our experiments. 
\subsection{Datasets}

We use PROX dataset~\cite{hassan2019resolving} for training where the ground-truth SMPL-X\cite{pavlakos2019expressive} parameters are generated by a fitting algorithm, which we denote as \textbf{pseudo-ground truth (p-gt)}. For training CVAE body synthesis network, we sample 4.3k bodies from 8 scenes as the training data. For training motion synthesis networks, we sample 10k two-second frames whose distance between start and end are larger than 0.5 meters in 8 scenes as the training data. For evaluation, we first test our short-term motion synthesis networks using 3k sequences in 4 unseen scenes using standard metrics introduced in the next section. We then perform human evaluation for which we generate 50 motion sequences each last for 2s, 4s and 6s. We also investigate the generalization ability of the models trained with the PROX dataset by directly testing them on the MP3D dataset~\cite{chang2017matterport3d}.

\begin{table*}
\begin{minipage}{0.56\linewidth}
\centering
\tablestyle{4pt}{1.0}
\footnotesize
\begin{tabular}{l|c|ccc|cc}
\multicolumn{1}{l|}{method} & \multicolumn{1}{c|}{scene} &  transl & orientation & pose & MPJPE & MPVPE  \\
\shline
RouteNet(t)+PoseNet & $\times$ &7.54 & 11.65 & 46.45 & 212.5 & 201.5 \\
RouteNet(t)+PoseNet & $\checkmark$ &7.31 & 11.82 & 46.73 & 213.1 & 202.1 \\

RouteNet(t+r)+PoseNet & $\times$ &7.58 & \textbf{9.38} & 44.16 & 201.6 & 190.5 \\

RouteNet(t+r)+PoseNet & $\checkmark$ &\textbf{6.91} & 9.71 & 44.89 & 195.8 & 184.9 \\

BaseNet & $\times$ & 9.33	& 10.61 & 44.09 & 235.8 &227.1  \\
BaseNet & \checkmark & 9.02	& 10.68 & 43.69 & 226.2 & 217.6  \\
Route+CVAE~\cite{zhang2020generating} & \checkmark & 8.47	& 10.03 & 59.16 & 294.2	 & 278.2	\\
HSNMS~\cite{xu2020hierarchical} & $\times$ & 10.01	& 13.72 & 63.41 & 293.5	 &275.2	\\
Ours w/o opt & $\checkmark$  & \textbf{6.91} & 9.71 & \textbf{41.17} & \textbf{191.6} & \textbf{180.9}  \\
\hline
Route+CVAE~\cite{zhang2020generating} w/ opt & $\checkmark$  & 9.60 & 9.97 & 61.65	& 311.8 & 293.8 \\
HSNMS~\cite{xu2020hierarchical} w/ opt &$\checkmark$  & 10.39 & 14.32 & 69.62	& 285.8 & 270.0 \\
Ours & $\checkmark$  & \textbf{8.06} & \textbf{9.53} & \textbf{46.68}	& \textbf{219.1} & \textbf{205.4}\\

\end{tabular}
\hspace{0.1cm}
\vspace{-0.3cm}
\caption{\textbf{Results of reconstruction error.} This table shows the comparison in reconstruction with baselines. The best results are shown in boldface.  $\checkmark$ in scene means this method has scene information and $\times$ means not. w/ opt means this method includes the optimization process and w/o means not.}
\label{tab:reconstruction error}
\end{minipage}
\ \
\begin{minipage}{0.42\linewidth}

\centering

\vspace{-0.60cm}

\tablestyle{4pt}{1.0}
\footnotesize
\begin{tabular}{l|cc|cc}
 \multirow{2}{*}{method} &\multicolumn{2}{c|}{non-collision} & \multicolumn{2}{c}{contact} \\
                 & PROX               & MP3D               & PROX            & MP3D            \\ \shline
Route+CVAE        &        98.57            &       95.19             &       88.68          &      91.36           \\
HSNMS~\cite{xu2020hierarchical}              &       96.88             &          95.29          &         99.33        &      90.12           \\
Ours w/o opt      &         97.53           &       95.33             &         99.32        &     92.59           \\
Route+CVAE~\cite{zhang2020generating} w/ opt &      99.88              &           99.51         &    99.17             &        95.06         \\
HSNMS~\cite{xu2020hierarchical} w/ opt       &        99.88            &                \textbf{99.57}   &   99.22              &      95.53           \\
Ours              &     \textbf{99.91}               &           99.30         &        \textbf{99.35}         &       \textbf{ 97.53 }         \\ \hline
p-gt                &     98.08               &         -           &       99.98          &      -           \\
\end{tabular}
\caption{\textbf{Results of the in-environment evaluation.} We use the non-collision score and modified contact score to evaluate how reasonable the generated motion is in the given scene. The best results are shown in boldface. It can be seen our method performs nearly the best in different scenes and our optimization process can help to improve the environmental adaptability of the baselines.}
\label{tab:env}
\end{minipage}
\vspace{-0.17in}
\end{table*}

\subsection{Evaluation Metrics}

\textbf{Reconstruction error metrics.} We introduce the standard metrics for evaluating the short-term motion: given the p-gt start and end human bodies, we aim to generate motion in between close to the p-gt. We first compare the reconstruction error by computing the $\ell_{1}$ distance (reported in $\times 100$) on translation $t$, orientation $r$, pose parameters $p$ between the predictions and the p-gt. We also use MPJPE~\cite{ionescu2013human3} and Mean Per Vertices  Position Error (MPVPE) both in millimeters to measure the mean distance between the predictions and the p-gt. In addition, start/end sides of the generated short-term motion sequence should be close to the input start/end bodies: We calculate the  $\ell_{2}$ distance between the input and generated start/end bodies. We refer this metric as \textbf{neighbour v2v distance}. Reducing this error pushes the end of one short motion sequence close to the start of the next short motion, which leads to better connections between two short motions for long-term motion synthesis.

\textbf{Naturalness metrics.} We perform naturalness evaluation for both short and long-term motion in two aspects: in-environment evaluation and human evaluation. For in-environment evaluation, we use the non-collision score~\cite{zhang2020generating} and contact score. We use Amazon Mechanical Turk(AMT) for human evaluation. For each task in AMT, we give a group of comparisons given the same inputs and ask users to score from 1 to 5 and a higher score means more natural. 
\vspace{-0.5em}
\subsection{Baselines}

We compare our approach with state-of-the-art baselines as well as the ablative variants of our method. Note previous motion synthesis works lack environment information and contact insensitive representation. For affordance prediction, previous approaches focus more on static/one-frame body generation. Thus, instead of directly comparing with previous approaches, we improve them to create stronger baselines. All baselines use the same modified SMPL-X representation. We introduce each baseline as following. 

\textbf{Route+CVAE.} Zhang et al.~\cite{zhang2020generating} propose to use a CVAE to synthesize a single body. However, this approach does not consider the motion information. Thus we provide the route using our RouteNet for CVAE to generate a sequence of human bodies. Specifically, we provide $\{t,r\}$ from RouteNet and adding conditions of $v_{s}$, $\beta$, $t$ and $r$ to CVAE. We call this baseline Route+CVAE. We apply optimization to it to improve the continuity.

\textbf{HSNMS}~\cite{xu2020hierarchical} is a state-of-the-art motion synthesis method combining both advantages of data-driven and neural network. For short-term motion, it searches the most similar motion sequence from a memory bank. 
Since this method lacks consideration of the environment, we also apply our optimization process to improve this approach.

\textbf{Ablative baselines.}
We compare with ablative baselines focusing on reconstruction error. We try to prove that PoseNet would perform better when having the predicted route as input so we design RouteNet+PoseNet which means the PoseNet separately generates the pose without given predicted route. We want to explore whether the rotation should be predicted in the route or pose network. RouteNet(t) means it only predicts $t$ and RouteNet(t+r) means it predicts $t$ and $r$; other parameters are generated in PoseNet. We also try an end-to-end neural network to predict route and pose jointly at one time and we name this network BaseNet. We also explore the effectiveness of scene information, thus for each ablative baseline, there is a version without scene information as inputs.

\vspace{-0.2em}
\subsection{Evaluation on reconstruction error}

\textbf{Comparison with baselines.} As shown in Tab.~\ref{tab:reconstruction error}, our method performs better in the accuracy of synthesis both in locations and poses than the baselines. Though the optimization process would slightly increase the error, it would largely improve the naturalness performance (Fig.~\ref{fig:optimization}). It can be seen our method has the lowest neighbour v2v distance from Tab.~\ref{tab:start_end}, which is easier to connect to long-term motion. 

\textbf{Comparison with ablative baselines.} As shown in Tab.~\ref{tab:reconstruction error}, we prove that predicted route as inputs would improve the performance of PoseNet and it is more accurate if we use RouteNet to predict $t$ and $r$. According to the results of BaseNet, directly using one network is not as good as our architecture. Finally, all the results with scene information would improve the accuracy of synthesis.

\begin{table*}[h]

\hspace{-0.25cm}
\begin{minipage}[t]{0.68\linewidth}
        
\centering

\tablestyle{4pt}{1.0}
\footnotesize
\begin{tabular}{l|cc|cc|cc}
 \multirow{2}{*}{method} &\multicolumn{2}{c|}{2 seconds} & \multicolumn{2}{c|}{4 seconds} & \multicolumn{2}{c}{6 seconds}
\\ 
     & PROX          & MP3D          & PROX          & MP3D          & PROX          & MP3D          \\ \shline
Route+CVAE~\cite{zhang2020generating} w/ opt &      2.23$\pm$1.10      &     3.06$\pm$1.07          &    2.75$\pm$0.98           &      2.94$\pm$0.99         &          2.83$\pm$1.06     &       3.19$\pm$0.95        \\
HSNMS~\cite{xu2020hierarchical} w/ opt &     2.90$\pm$1.09          &       3.21$\pm$0.97        &     3.26$\pm$1.02          &      3.14$\pm$0.99         &      3.22$\pm$1.08         &      3.13$\pm$1.05         \\
Ours &      \textbf{3.39$\pm$1.11}         &      \textbf{3.33$\pm$0.90}         &           \textbf{3.55$\pm$0.88}    &      \textbf{3.22$\pm$1.05}         &       \textbf{3.65$\pm$0.96}        &       \textbf{3.30$\pm$0.96}        \\ \hline
p-gt   &      3.60$\pm$1.10         &      -         &     3.76$\pm$0.97          &      -        &        3.92$\pm$0.91       &      -        \\

\end{tabular}
\hspace{-0.3cm}
\vspace{-0.1in}
\caption{\textbf{Results of the human evaluation.} We show human evaluation for motions last for 2 seconds, 4 seconds and 6 seconds. We provide the average human evaluated score(1-5) w.r.t. the average $\pm$ the standard deviation. The best results are shown in boldface.}
\label{tab:human}
\end{minipage}
\quad
        \begin{minipage}[t]{0.30\linewidth}

\centering


\tablestyle{4pt}{1.0}
\footnotesize

\begin{tabular}{l|cc}
 \multirow{2}{*}{method} &\multicolumn{2}{c}{neighbour v2v distance}  \\
                 & w/ opt               & w/o opt                          \\ \shline

Route+CVAE~\cite{zhang2020generating}               & 21.78             & 10.59            \\
HSNMS~\cite{xu2020hierarchical}                    & 21.05             & 10.84            \\
Ours                    & \textbf{8.25}             &        \textbf{7.31}           \\ \hline
p-gt                    &     3.59         &        -           \\
\end{tabular}
\hspace{0.3cm}
\vspace{-0.1in}
\caption{\textbf{Neighbour v2v distance with and without optimization.} The best results are shown in boldface.}
\label{tab:start_end}

\end{minipage}
\vspace{-0.3in}
\end{table*}

\subsection{Evaluation on naturalness} 
\begin{figure}[]
\vspace{0.1in}
\centering
\includegraphics[width=0.9\linewidth]{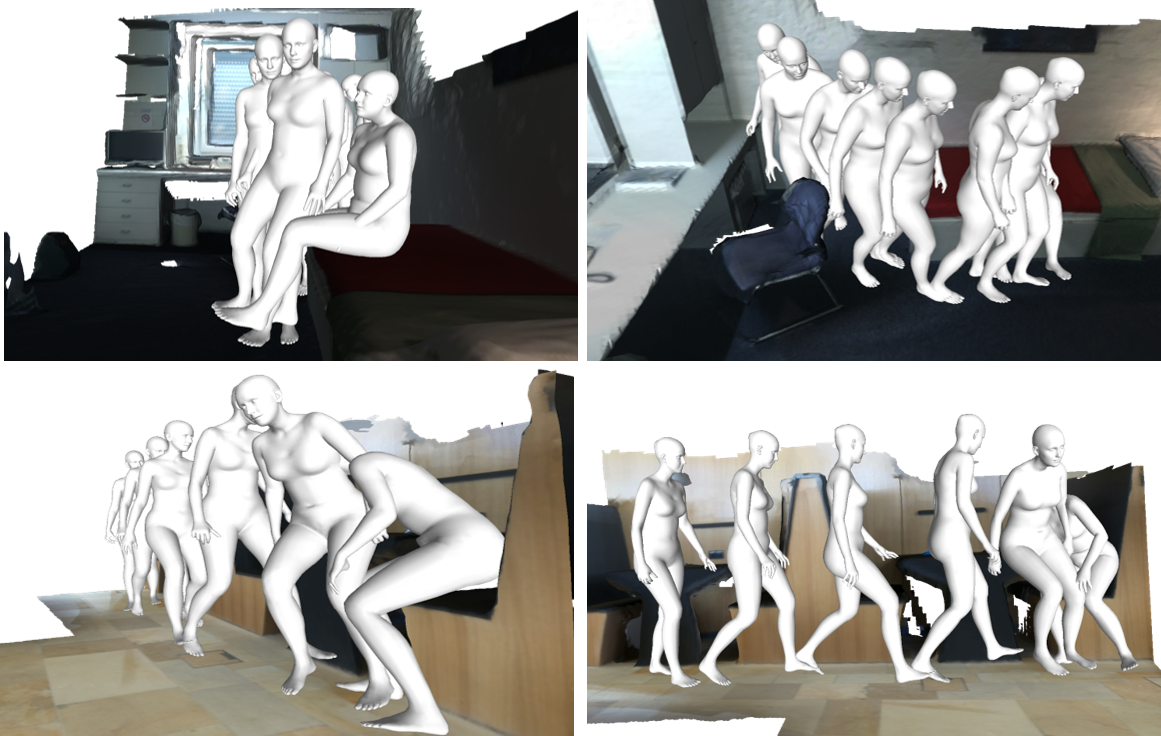}
\vspace{-0.1in}
\caption{Two pairs of results in two views in PROX~\cite{hassan2019resolving} scenes. Both rows show a human walking and then sitting down.}
\vspace{-0.25in}
\label{fig:two_view}
\end{figure}

We firstly show qualitative naturalness comparison for 2-second motion in Fig.~\ref{fig:2_seconds_compare}, and 4-second motion in Fig.~\ref{fig:4_seconds_compare}. Our method can produce motions close to the real record, with less unreasonable motion and better human-scene interaction. We show two pairs of our results in two views in Fig.~\ref{fig:two_view}. We also perform in-environment naturalness evaluation between these methods. As shown in Tab.~\ref{tab:env}, our method has nearly the best performance in both contact and avoiding the collision. Not only that, our optimization process can largely improve baselines' performance on generating physically plausible motion. 

For human evaluation, we compare our approach with the other two baselines and the p-gt from PROX~\cite{hassan2019resolving} dataset. For a fair comparison, the start, end and sub-goal bodies are from PROX~\cite{hassan2019resolving}. Since the p-gt is generated by fitting, it would also have implausible motion sequences. As shown in Tab.~\ref{tab:human}, our method has the highest score compared to the other two baselines for 2s, 4s and 6s motion and the scores are close to the p-gt, which shows our methods can generate realistic and natural motions. Through the experiments, we prove that compared to previous affordance learning method, our method has a better performance in motion continuity. Previous motion-based method which heavily relies on existing data may have problems in handling the environment. 
\label{sec:naturalness}

\begin{figure}[]
\centering
\vspace{0.1in}
\includegraphics[width=0.9\linewidth]{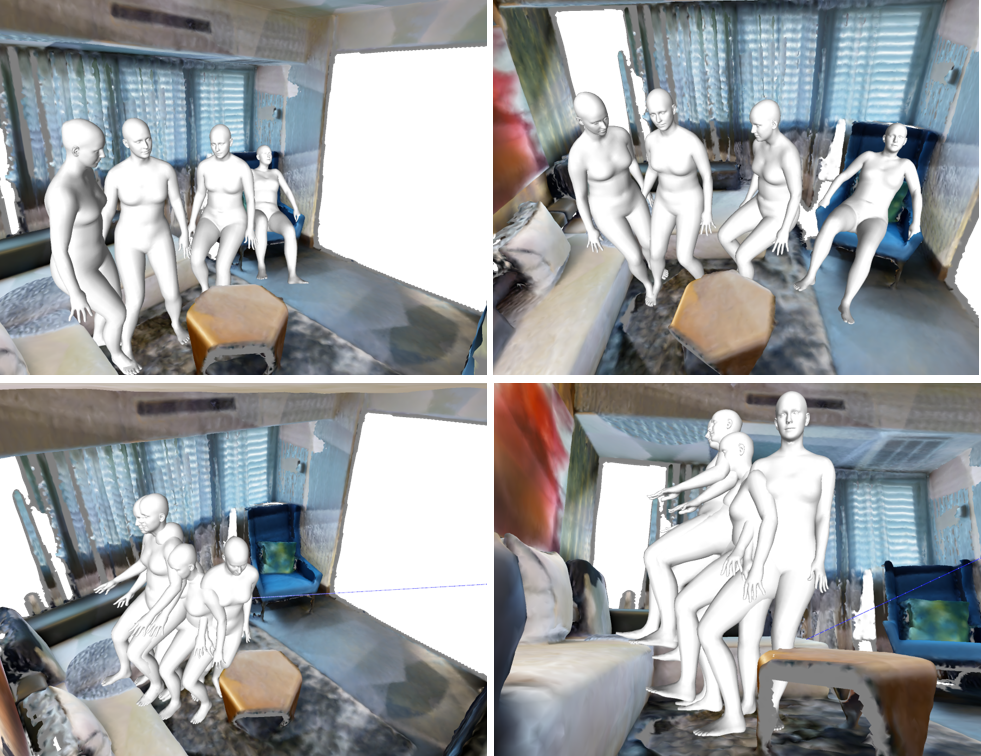}
\vspace{-0.08in}
   \caption{Two pairs of our results in two views in the scene from MP3D~\cite{chang2017matterport3d}. The first row shows a human standing up and walking and the second row shows a human jumping on the sofa.}
\label{fig:two_view_mp3d}
\vspace{-0.05in}
\end{figure}

\textbf{Generalization on MP3D~\cite{chang2017matterport3d} dataset.} We evaluate naturalness performance with our trained model directly on MP3D. We randomly sample the input shape, positions and orientations for the start/end and sub-goals and generate plausible bodies using our body synthesis model. We apply our motion synthesis networks to the input bodies and then optimize the whole motion sequence. The input start/end and sub-goal bodies for baselines are the same. We show in-environment evaluation in Tab.~\ref{tab:env} and human evaluation in Tab.~\ref{tab:human}. Since the positions and orientations are randomly selected, there would exist some challenging cases, such as jumping on the sofa. We also provide our results shown in two views in Fig.~\ref{fig:two_view_mp3d}. Generally, our method can generate the most realistic results with the best environment adaptability. 

\begin{figure}[]
\vspace{-0.06in}
\centering
\includegraphics[width=0.95\linewidth]{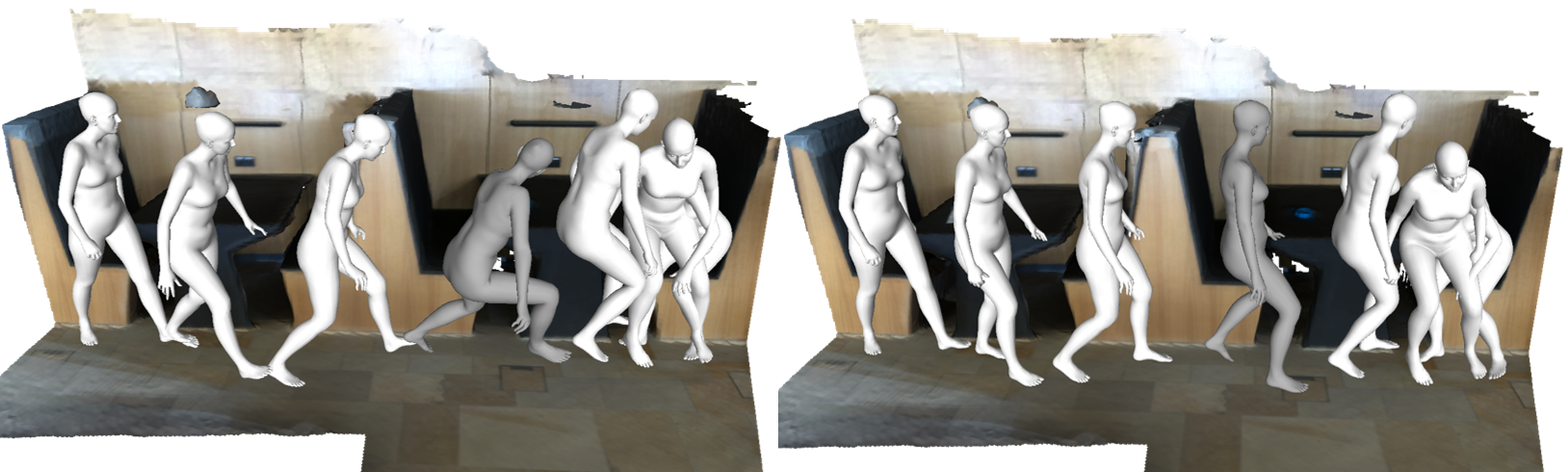}
\vspace{-0.07in}
   \caption{We show examples of motion with different generated sub-goal bodies (gray color bodies).}
\label{fig:subgoal}
\vspace{-0.2in}
\end{figure}

\textbf{Diversity.} One advantage of our framework is we can control the shape, positions and gestures of the sub-goal bodies (including start/end), which makes the motion more diverse. We show an example in Fig.~\ref{fig:subgoal}, since we can generate different sub-goal bodies, the motion sequence would look differently. This could enhance the interaction between human and scene and generate diverse motions.

\vspace{-0.2em}

\vspace{-0.1em}
\section{Conclusion}
\vspace{-0.1em}
In this paper, we propose a novel hierarchical generative framework to synthesize long-term human motion in the 3D scene. We generate the sub-goal bodies and short-term motions with two deep models both designed with human-scene interaction. We further design an optimization-based method to improve realistic synthesis and connect the short-term motions to a long-term motion. Compared with other baselines, our framework can synthesize more natural and physically plausible long-term motion in the 3D scene.

{\footnotesize \textbf{Acknowledgements.}~This work was supported, in part, by grants from DARPA LwLL, NSF 1730158 CI-New: Cognitive Hardware and Software Ecosystem Community Infrastructure (CHASE-CI), NSF ACI-1541349 CC*DNI Pacific Research Platform, and gifts from Qualcomm and TuSimple.}
{\small
\bibliographystyle{ieee_fullname}
\bibliography{egbib}
}

\clearpage
\setcounter{section}{0}

\begin{center}
\textbf{\Large Appendix}
\end{center}

We provide more details about datasets, optimization implementation, discussion of other methods, and naturalness evaluation in the appendix.

\section{Datasets}

We use  `MPH112', `MPH11', `MPH8', `N0Sofa', `N3Library', `N3Office', `BasementSittingBooth' and `Werkraum' in PROX\cite{hassan2019resolving} as training scenes and we use `MPH16', `MPH1Library', `N0SittingBooth', `N3OpenArea' in PROX\cite{hassan2019resolving} and the family room, living room and bedroom of `17DRP5sb8fy' in MP3D\cite{chang2017matterport3d} dataset as testing scenes.

For the training data of sub-goal body synthesis network, we down-sample the original motion sequences and use the static body every 0.33 seconds. For the training data of motion synthesis networks, we first sample the start and end bodies which has a duration of 2 seconds and the Euclidean distance between them is larger than 0.5 meters. We use the motion in between these start/end pairs as our motion training data.

\section{Implementation details}
\label{sec:apend:detail}
To better balance the environmental constraints and plausibility of motion, we perform our optimization in two stages. In the first stage, we enhance the optimization for environment constrains and motion smoothness and set $\lambda_{foot}=0$, $\lambda_{col}=1$, $\lambda_{cont}=1$ and $\lambda_{smooth}=0.25$. In the second stage, we want to improve the motion plausibility and set $\lambda_{foot}=1$, $\lambda_{col}=1$, $\lambda_{cont}=1$ and $\lambda_{smooth}=0.25$.

\section{Discussion of other methods}
\begin{figure}[]
\begin{center}

\includegraphics[width=0.9\linewidth]{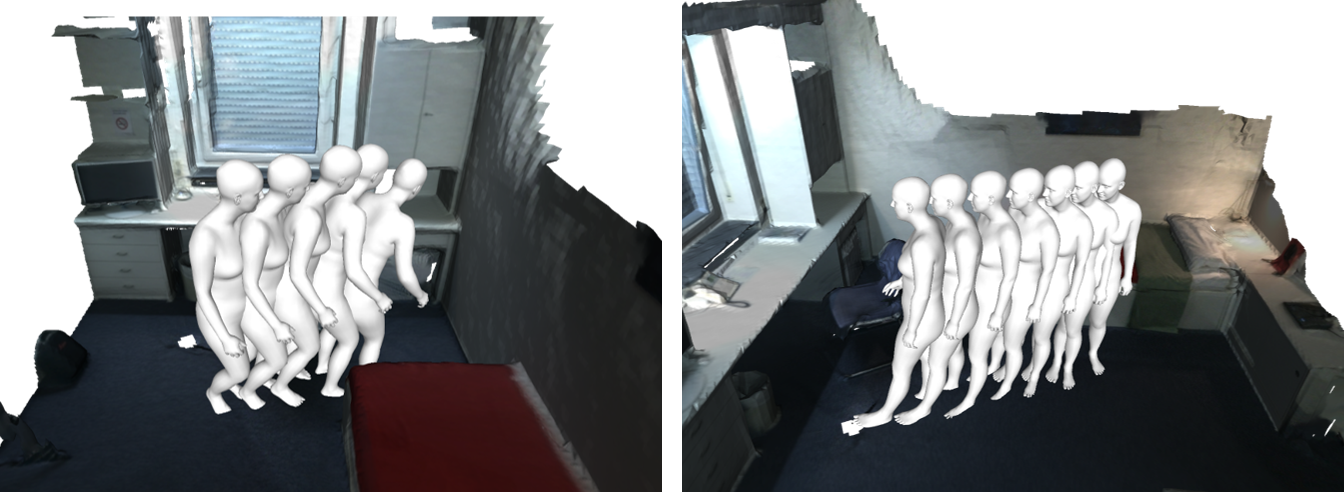}
\end{center}

\vspace{-0.15in}
   \caption{Two examples of CVAE interpolation results. Left example shows that if the start and end bodies we give are without legs walking motion, the result of interpolation is more like standing but being pushed forward. The right example shows that if there is legs changing motion but with a certain distance, no matter how far the motion is, the interpolation result will finish this in one step, thus the whole motion is full of foot skating. }
\vspace{-0.2in}
\label{fig:interpolation}
\end{figure}

We also try to create a baseline inspired by CVAE interpolation for motion synthesis. Since our setting is to give the start and end bodies to generate motion in between, we first perform gradient descent with Adam~\cite{kingma2014adam} to fit two latent z of the start and end bodies. After we get the latent z of the start/end, we can use interpolation to get the sequence in between. However, this method may only be applied to a few cases. For motion with a certain distance, this method is more like average interpolation rather than following the law of human motion. As shown in Fig.~\ref{fig:interpolation}, 
CVAE interpolation can not generate a complete human motion.

Another related work is~\cite{cao2020long}, which uses past 1 second motion to predict future 2 seconds motion using skeleton and rgb images to represent human and scenes. Their motion is in 10fps and ours is 30fps. Their paper's w/ gt destination setting is the most similar to ours. They report the path error and MPJPE~\cite{ionescu2013human3} error in PROX~\cite{hassan2019resolving} which can be compared to us. Their path error is from 19.3 to 23.7, and the average of ours is 8.06. For MPJPE in millimeters, ours is 219.1 while their method is 237. Considering their weakness in generating dense motion sequence, large requirements of past sequence and different representation setting, we do not compare the other aspects.

\section{Naturalness evaluation}
\label{sec:apend:nature}
We provide more details about modified contact score, human evaluation and show more qualitative results in this section.

\textbf{Modified contact score.} Since our task is a motion synthesis task, we set a threshold of 0.01 of the signed distance value and if it is smaller than 0.01, we take it as contact, unlike 0 in \cite{zhang2020generating}.

\textbf{Human evaluation details.} Different from \cite{3d-affordance} giving two examples once and asking user to compare which is better and \cite{zhang2020generating} giving just one example to score from 1 to 5, we give 4 examples (two baselines, ours and pseudo-ground truth) once with the same start, end, sub-goal body inputs and ask users to score from 1 (strongly not natural) to 5 (strongly natural) each.  The advantage of this is we can ensure that for the same motion, people who scored are the same, which is fairer for the comparison. Each task will be scored by 3 users and we calculate the average score.

\begin{figure*}[]
\begin{center}

\vspace{0.5in}
\hspace{0.5in}

\includegraphics[width=1\linewidth]{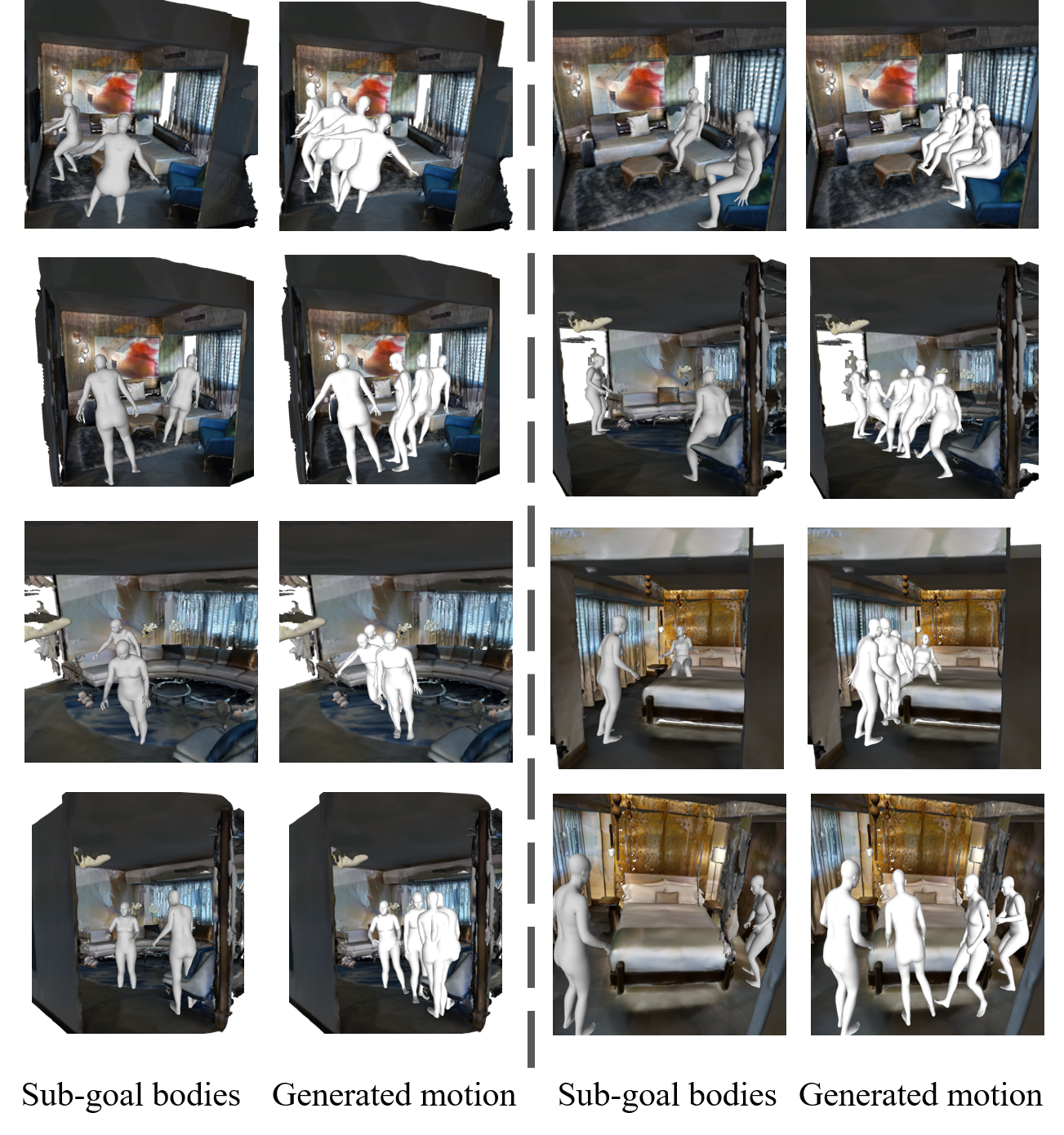}
\end{center}
   \caption{\textbf{Our results.} We show the generated sub-goal bodies and motion between in sub-goal bodies.}
\vspace{0.5in}
\label{fig:2seconds}
\end{figure*}

\begin{figure*}[]
\begin{center}
\vspace{0.5in}
\hspace{1in}

\includegraphics[width=1\linewidth]{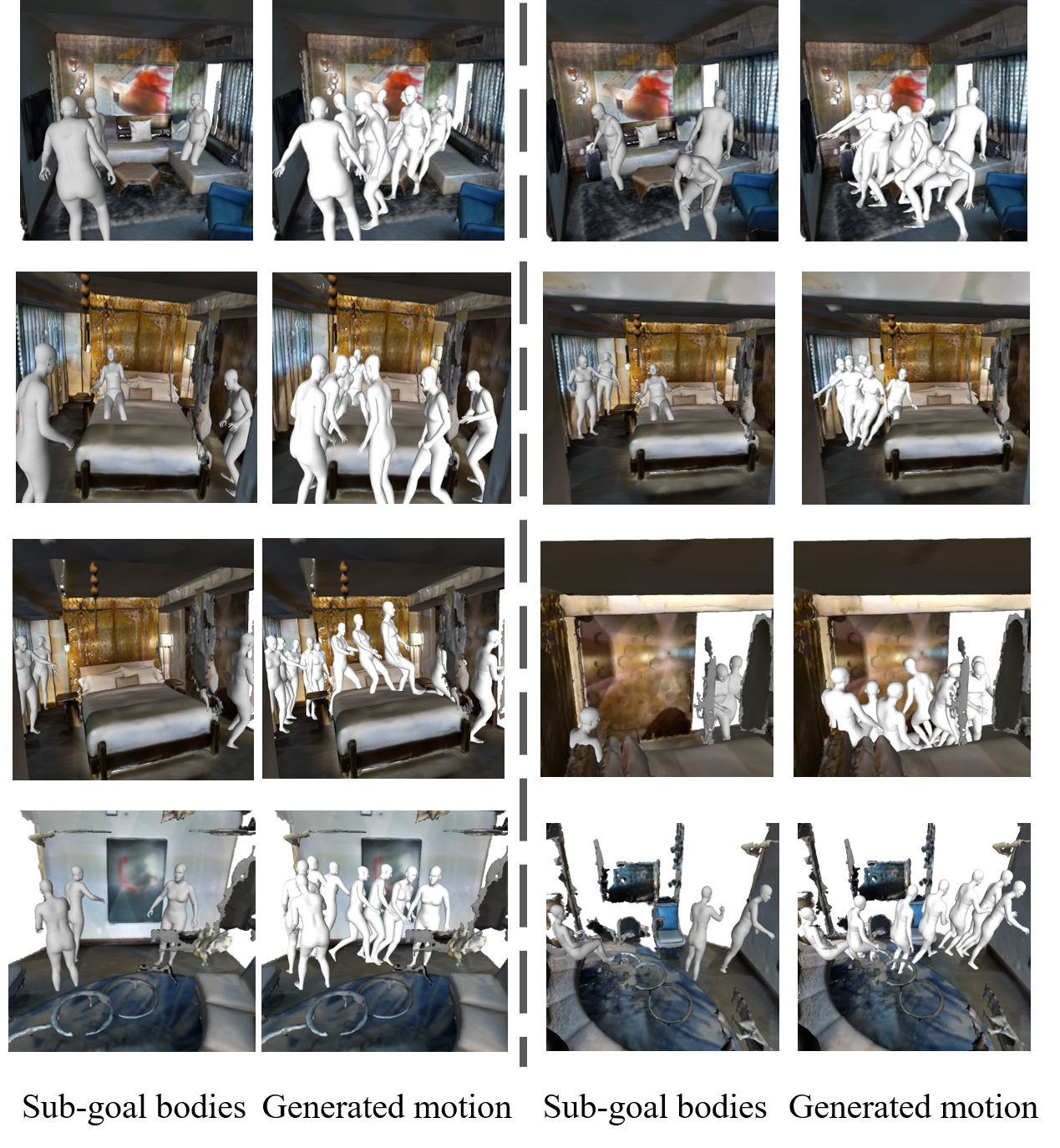}
\end{center}
   \caption{\textbf{Our results.} We show the generated sub-goal bodies and motion between in sub-goal bodies.}
\vspace{0.5in}
\label{fig:4seconds}
\end{figure*}

\begin{figure*}[]
\begin{center}
\vspace{-0.25in}

\includegraphics[width=0.9\linewidth]{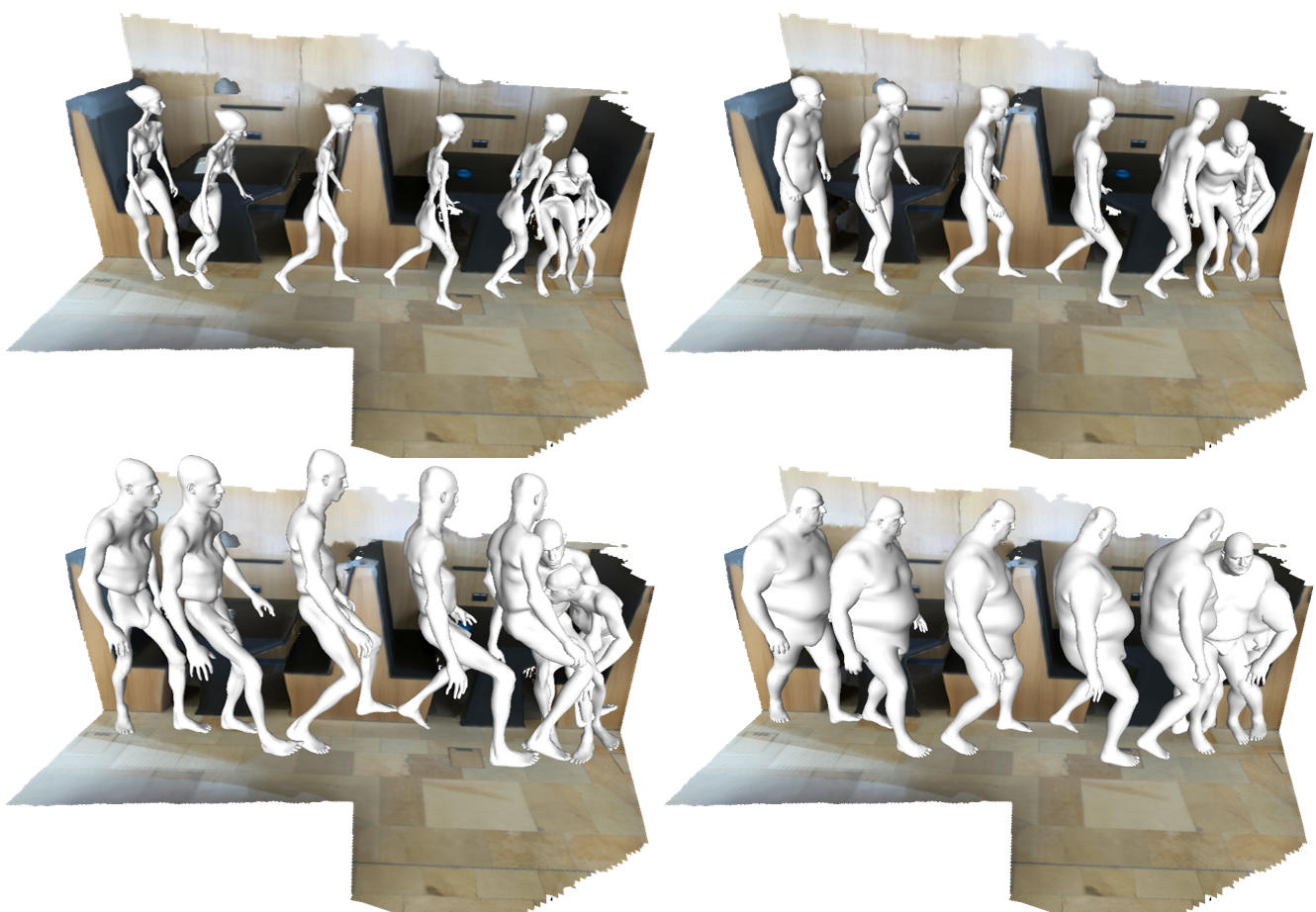}
\end{center}
\vspace{-0.09in}
   \caption{Four examples of diversified motion with different body shape $\beta$. }

\label{fig:diversity shape}
\end{figure*}

\begin{figure*}[]
\begin{center}
\vspace{-0.2in}

\includegraphics[width=0.9\linewidth]{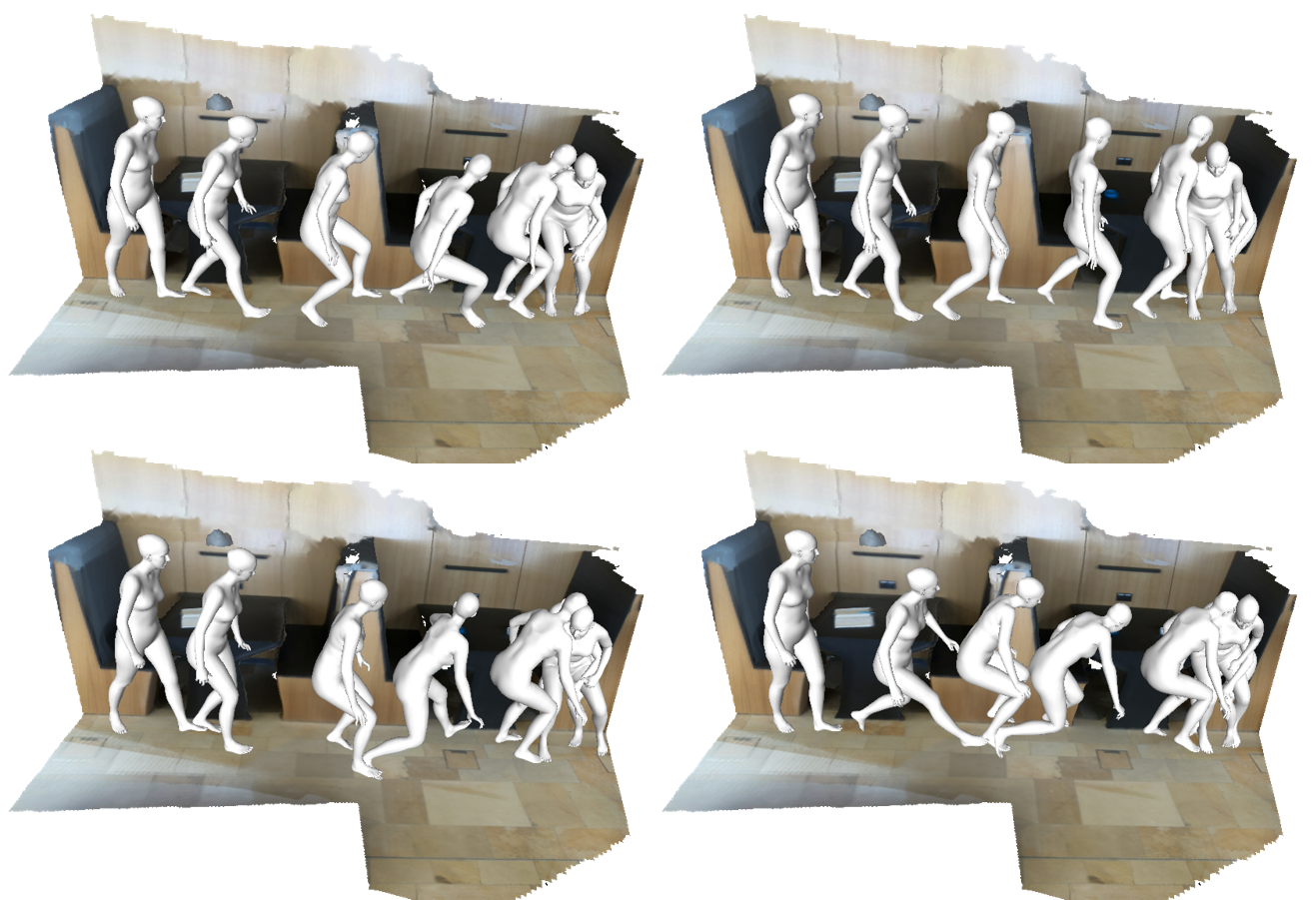}
\end{center}
\vspace{-0.09in}
   \caption{Four examples of diversified motion with different latent z for sub-goal body. }

\label{fig:diversity z}
\end{figure*}

\textbf{More qualitative results.} We provide more qualitative results of our generated sub-goal (start/end) bodies and generated motion in between in different scenes in Fig.~\ref{fig:2seconds} and Fig.~\ref{fig:4seconds}. It can be seen that our method can synthesize different kinds of challenging long-term motion such as walking, sitting down, jumping on the bed and lying down in different scenes. Furthermore, we provide examples of randomly sampled body shape $\beta$ in Fig.~\ref{fig:diversity shape} and also examples of randomly sampled latent variables for sub-goal bodies in Fig.~\ref{fig:diversity z}. It can be seen that our method can synthesize diversified motion with different body shape and different motion style.

\end{document}